%% file: main.tex
\documentclass[transmag]{IEEEtran}
\usepackage{latexsym}
\usepackage{amsfonts,amssymb,amsmath}

\usepackage{booktabs}
\usepackage{multirow}
\usepackage{fixmath}
\usepackage{subcaption}
\usepackage{float}
\usepackage{xspace}
\usepackage{hyperref}
\usepackage{cite}
\usepackage{xcolor}

\usepackage{graphicx}
\usepackage{bmpsize}
\usepackage{grffile}

\def\BibTeX{{\rm B\kern-.05em{\sc i\kern-.025em b}\kern-.08em T\kern-.1667em\lower.7ex\hbox{E}\kern-.125emX}}
\begin{document}

\title{Disentangling private classes through regularization}

\author{Enzo Tartaglione, \IEEEmembership{Member, IEEE}, Francesca Gennari, and Marco Grangetto, \IEEEmembership{Senior Member, IEEE}
\thanks{E. T. is with 
the LTCI, Telecom Paris, Institut Polytechnique de Paris (e-mail: \{enzo.tartaglione\}@telecom-paris.fr), F. G. Author, is with LAST-JD-RIoE consortium, more specifically Mykolas Romeris University - University of Bologna - University of Turin 
(e-mail: francesca.gennari8@unibo.it), M. G. is with 
the Computer Science Department at University of Turin, Torino, Italy (e-mail: \{marco.grangetto\}@unito.it).}
}

\IEEEtitleabstractindextext{

\begin{abstract}
Deep learning models are nowadays broadly deployed to solve an incredibly large variety of tasks. However, little attention has been devoted to connected legal aspects. In 2016, the European Union  approved the General Data Protection Regulation which entered into force in 2018. Its main rationale was to protect the privacy and data protection of its citizens by the way of operating of the so-called ``Data Economy''. As data is the fuel of modern Artificial Intelligence, it is argued that the GDPR can be partly applicable to a series of algorithmic decision making tasks before a more structured AI Regulation enters into force. In the meantime, AI should not allow undesired information leakage deviating from the purpose for which is created. In this work we propose DisP, an approach for deep learning models disentangling the information related to some classes we desire to keep private, from the data processed by AI. In particular, DisP is a regularization strategy de-correlating the features belonging to the same private class at training time, hiding the information of private classes membership. Our experiments on state-of-the-art deep learning models show the effectiveness of DisP, minimizing the risk of extraction for the classes we desire to keep private.
\end{abstract}

\begin{IEEEkeywords}
Deep learning, GDPR, regularization, disentangling, artificial neural networks, private classes.
\end{IEEEkeywords}

}

\maketitle

\input{sections/1_introduction}
\input{sections/2_method}
\input{sections/3_attacks}
\input{sections/4_biasedmnist}
\input{sections/5_experiments}
\input{sections/6_conclusion}


\section*{Acknowledgment}

This project has received funding from the European Union's Horizon 2020 research and innovation programme under the Marie Skłodowska-Curie ITN EJD grant agreement No 814177.

\bibliographystyle{IEEEtran}
\bibliography{main}



\end{document}

%% file: sections/1_introduction.tex
\section{Introduction}

Currently, a significantly large portion of problems is being solved through the deployment of deep learning model, considered by most as a ``universal problem solving tool''~\cite{sonoda2017neural}. Their fast and uncontrolled spread, however, requires legal rules that are influencing or are bound to influence the development of Artificial Intelligence (AI). There are two main reasons for this.
Firstly, it is of general interest that the algorithmic decision making systems that are being described and explained further can be of a wide application, irrespective of the private
or public entity of the user. There are tools, available in the literature, attempting to provide explanations on the AI's decision process, like GradCam~\cite{selvaraju2017grad}. However, these tools are limited to input-output relationships, and do not really provide insights on how to really open ``the black box''. Secondly, it would be irresponsible and unethical to promote algorithms that could possibly create damages to the privacy of people whose data are used by the AI~\cite{Osia2020AHD}. To this end, what will be succinctly explained is the legal outlook of the European Union on AI. This is done because the EU has already obtained a role of legal influencer as far as the discipline of data protection is
concerned by drafting the General Data Protection Regulation (GDPR)~\cite{art1} and because
it has started since 2018 in trying to create the conditions to regulate the AI. It all
started with the Ethical Guidelines of the AI~\cite{art2}, the Assessment list for trustworthy
Artificial Intelligence (ALTAI) for self-assessment~\cite{art3} and finally the proposal for the
regulation of the AI~\cite{art4}. These two issues will be dealt with subsequently. It is important
to specify that the following two subsections are not a commentary of the entirety of the
two regulations but a focus on the most problematic aspects which could potentially
clash with the development of an AI described further.

\begin{figure}
    \centering
    \includegraphics[width=\columnwidth]{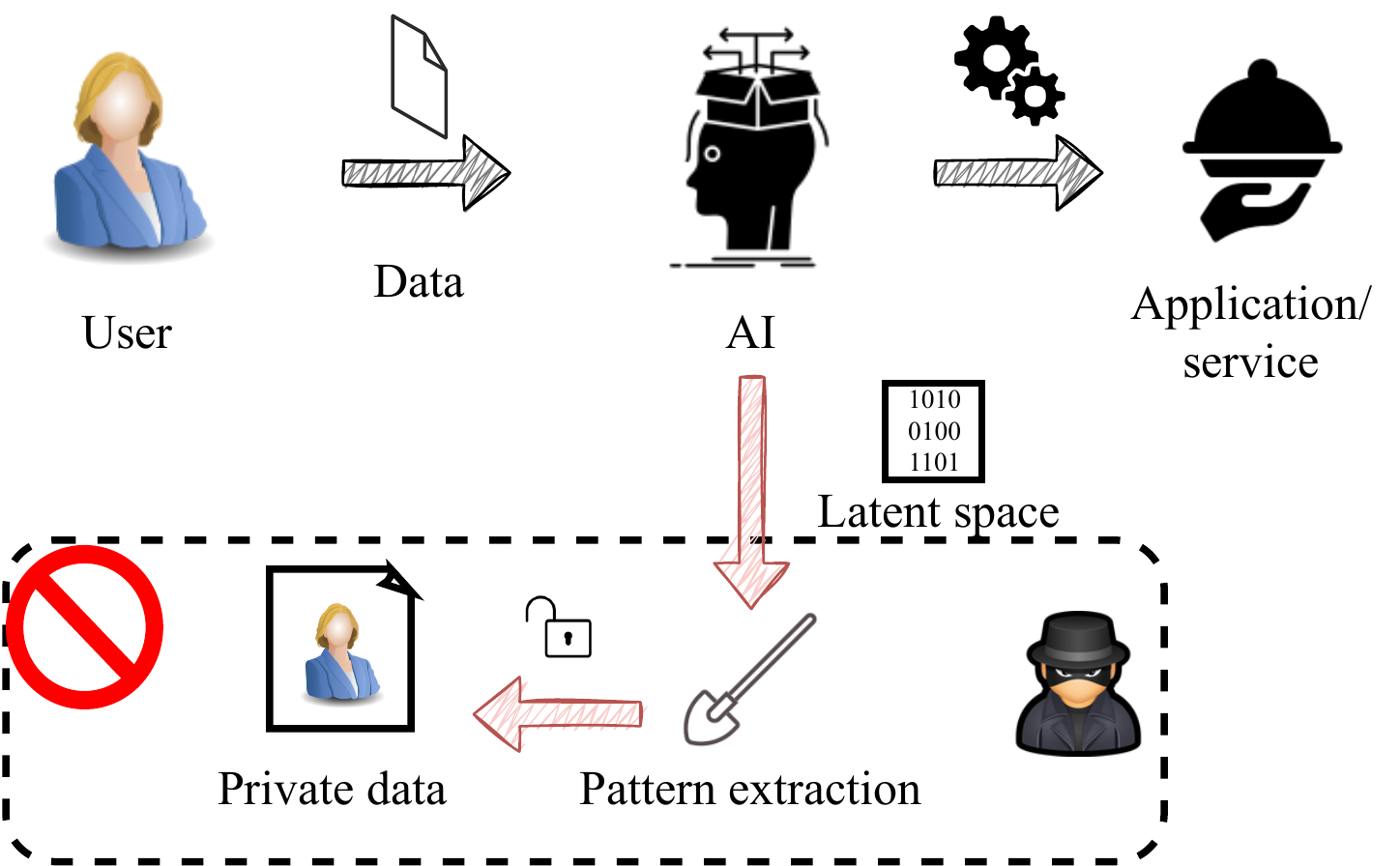}
    \caption{From the latent space of ANN models there might be information leakage allowing an attacker to recover sensitive/private data. Our goal is to prevent this from happening hiding this private information.}
    \label{fig:teaser}
\end{figure}

\subsection{The GDPR and the AI}
The GDPR is the first regulation (a EU mandatory legislative act) that tries to balance the right to data protection, and, more largely, to privacy in the EU, with the Big Data economy. The points of contact between AI developers and the GDPR are mainly two: the concept of personal data (i) and of data processing (ii). Then it is important to understand how the developers of neural networks and/or companies/research institutes are seen according to the GDPR (iii) and how the principles of data protection by design and by default (iv) could be relevant for the development of trustworthy and ethical AI models.\\

\textbf{Personal data.} Whenever one works with data that can be referred to people (and the AI creation field makes no exception), lawyers but especially innovators must follow the GDPR not only to check on compliance but also to ensure that the AI does not infringe on people’s privacy and data protection, which in the EU are considered fundamental rights and therefore are always to be considered hierarchically superior to economic rights. 
Concerning data relating to a specific individual, Article 4.1 1) GDPR defines personal data as anything ‘relating to an identified or identifiable natural person’. Some scholars have warned about the dangers that an overly large definition of personal data might have on the applicability of the raison-d’être of this regulation~\cite{purtova2018law}. In more technical terms, personal data is information that can lead to the identification of an individual.\\

\textbf{Data processing.} Further, GDPR is relevant for the application of the AI model as its operations can be described as data processing when performed on personal data. Article 4.2 GDPR describes processing as practically any operation involving personal data. Therefore, whenever an AI model uses personal data (to train or to give a specific output autonomously), there will be a processing activity. The GDPR defines certain categories of more sensitive personal data at Article 9.1 GDPR. These include biometric, genetic and general ‘health data’ (the latter are better described in Article 4.1 15) GDPR). These could also be categories of data used by the model. Sensitive data in principle should not be processed but there are exceptions that can justify such operations. In Article 9.2 h) GDPR increasing the efficiency of a medical diagnosis (besides the expressed specific consent of the patient, see 9 2. a)) can be a justification for sensitive personal data processing. In this model the function of the data processing is not to identify people but to create a technology that averts discriminatory results and can increase the efficiency of medical diagnosis. This model would also fulfil the respect of the principle of fairness as set in Article 5 1. a) GDPR.\\

\textbf{The stakeholders: accountability and liability.} There are two main stakeholders that can be involved in the development of AI technologies and under a GDPR perspective.  The first one is the controller (Article 4.7 GDPR) which is the ‘natural, legal person or public authority…which, alone or jointly with others determines the purposes and the means of the processing of personal data’. In our particular case, the controller should be either the public or private actor that decides to use the model. However, the subject/entity which handles concretely the data processing phase is called processor (Article 4.8 GDPR) and could be a company to which these processing functions are delegated by the controller. The GDPR, by stating the principle of accountability (Article 5.2 GDPR) and of responsibility/liability of the controller (Article 24 GDPR and Article 82 GDPR), mandates that the same controller must implement the technical measures that are requested in order to ensure the security (Articles 32-35 GDPR) of the data subject (the person whose data are being processed). This means that the entities responsible for the training and development of neural networks will not be exempt from these rules whenever data processing activities involve personal data. Security and safety of personal data are expected by the controller as a concrete application of the principles of fairness, transparency and lawfulness (Article 5.1 a) GDPR).\\

\textbf{Data protection by default and by design as a tool.} The GDPR is a system with a regulatory and compliance function, but, at the same time, it is open-ended as it envisions the protection of personal data as a result and does not focus on how to achieve this target, as it depends also by the technical and economic means available in each specific case. That is why Article 25 GDPR sets out the principle of privacy by design and by default. It means that privacy should be the objective through which all the processing activities of a data driven technology must aim at from their very first conceptualisation~\cite{leenes2018artificial} and, also, privacy should be the default option, which means that it should revolve around the amount of data that is necessary for the processing and not to exceed that limit (this principle is called data minimisation principle and can be found at Article 5.1 c) GDPR)~\cite{voigt2017eu}. The general character of this article also leaves way not only to business and management methods, but also to technology itself the task to find ways that can increase the privacy and security of the data subject~\cite{bygrave2017data}. The model could be a concrete application of this principle as it protects the identity of the subject, hence their privacy, from the first conceptualisation to its concrete application.

\subsection{New perspectives}
The proposed regulation for AI by the European Commission adopts two main views on regulating this technology. On the one hand, the focus is on the accountability principle which makes the regulation less detailed than regulations are in general. The scope of that is to make it a piece of legislation that is technologically neutral as much as possible and to resist the evolution of this kind of technology through time. On the other hand, risk assessment is the rationale through which this regime works. There are three main kinds of AI according to the regulation. Firstly, there are the AI algorithms that are forbidden and consist of technologies that allow the surveillance and the subliminal persuasion of people and especially of minorities and identifiable groups of people. There are specific exceptions that are provided for public authorities’ surveillance functions.  
Secondly, there are the so called high risk AI technologies which for the moment are listed in Annex I of the proposal. They include all kinds of machine learning, logic and knowledge based approaches (including knowledge representation, inductive and logic programming, inferences and deductive engines, symbolic reasoning and expert), statistical approaches, Bayesian estimation, search and optimization methods. These technologies must be evaluated for their objective and the risk they might entail for EU citizens. To this extent there will be a EU database for the high risk AI algorithms which will be open and accessible for all EU citizens (Article 60 of the proposal). Each Member state will have to set up an authority (or to confer new competences to old ones) in order to check on the work of specific Notified Bodies (Title III, Chapter 4). These bodies will be private, semi-private or public organisations with the task of certifying that AI algorithms are state of the art and secure. There will be a European Artificial Intelligence Board (Title VI) which will be composed of the national agencies or authorities which will be in charge of the coordination and monitoring of the algorithms on the market. This proposal, if voted favourably by the European Parliament and the Council of the EU, will become biding in 27 states and there is some reason to believe that it will serve as a legal model for other countries in the future just as the GDPR did.

\subsection{Technological context}
Towards privacy preservation, cryptographic techniques cover a prominent role and have broad application. In particular, attribute-Based Encryption (ABE) is a promising cryptographic primitive which is able to implement access control for secure data storage, for example, in the cloud~\cite{horvath2015attribute, zhang2020attribute}, or even probabilistic data structures~\cite{singh2020probabilistic}. Besides, homomorphic-based encryption involves the necessity to have a secret key to access the data, and is one of the mostly used approach to data protection~\cite{acar2018survey}. Classical approaches like the above-mentioned, however, protect the data entirely, without leakage. Nowadays the users are more and more willing to share contents to get services, sharing even personal data in order to get a service. Standard cryptographic techniques are inapplicables in these contexts, especially considering that most of the data processing approaches involve the utilisation of Artificial Neural Networks (ANNs). Towards this end, the proper design of an algorithm, which is not entirely crypting the data but which filters a part of the information to be kept private, is necessary.\\
Privacy-aware learning is not a novel concept in machine learning. One of the very first works in such an area was published in the far 1965 by Warner~\cite{warner1965randomized}. In particular, they were suggested privacy-preserving methods for survey sampling. Following this path, in the 70s many works have been proposed on different areas, like census taking and analysis of tabular data by Fellegi~\cite{fellegi1972question}.\\
Very recently, thanks to the increase of computational capabilities, many works have been proposed on privacy-preserving in computational frameworks. A work by Dwork et al.~\cite{dwork2016calibrating} studied how much noise is required to guarantee ``differential privacy''~\cite{dwork2009differential} from data. Following these paths, a more recent work by Duchi et al.~\cite{duchi2014privacy} formalized convergence boundaries for training and the trade-off between privacy guarantees and the utility of the resulting statistical estimators. This knowledge has been also recently applied to deep learning frameworks, with a work by Abadi et al.~\cite{abadi2016deep}, by introducing some tuned noise in the update rule.\\
A very recent work by Chamikara et al.~\cite{chamikara2019efficient} proposes the insertion of a ``randomizing layer'' to guarantee anonymization in the features used for training. While this approach certainly guarantees no locally-sensitive information leak under some constraints, it is a very limiting approach as it requires the randomization to happen after convolutional layers. We know nowadays most of the ANNs have few fully-connected layers or they are even fully-convolutional, and the proposed approach is not applicable without the further insertion of fully-connected layers. Furthermore, it introduces computational overhead, and full back-propagation on the entire model is not suitable as no assumption on the random sampling can be done (so the entire convolutional part must be pre-trained in a non-private fashion and is not further modified).\\
A different approach to preserve data privacy is the so-called ``federated learning'' approach. In general, private datasets are held by the proprietary of the data, who is directly training a local neural network model. Now, the parameters of the models are sent to a master node, who is then propagating to all the private computational nodes the general configuration of the parameters. This approach has been proposed by Shokri and Shmatikov~\cite{shokri2015privacy}, and allows parallel and private computation. However, such an approach does not take into account any ethical bias, like gender or race: what it guarantees is that the original data are not directly shared, but some sensible information actually are.\\
In our framework, we train a deep neural network model on some target classes, but we desire at the same time to minimize the risk of information leakage for some classes we desire to keep private. Towards this end, we propose a regularization strategy to \underline{Dis}entangle \underline{P}rivate classes: we define a bottleneck layer $\Gamma$ in our model and we de-correlate features belonging to the same bias class but having different target. We have two contributions to DisP: a memory term, which keeps track of the average correlation in the full training set, and a local mini-batch term which orthogonalizes all the features belonging to the same bias class but different targets. To the best of our knowledge, DisP is the first approach in this context having generality towards the neural network architecture (or in other words, it is applicable to any deep learning model), providing insights on the latent space representation of the information, and practically showing the features' correlations. With this work, we move the first steps to ``open the black box'' ANN models are, and we accomplish this putting a constraint on the information bottleneck of our model, making comparisons with other approaches impossible with the same experimental setup. An high-level view of our framework is displayed in Fig.~\ref{fig:teaser}. \\
The rest of the paper is organized as follows. Sec.~\ref{sec:method} provides details about DisP working principles, Sec.~\ref{sec:attacks} presents two possible attacks to retrieve the private information, one unsupervised and another one supervised. Sec.~\ref{sec:control} introduces biased-MNIST, a dataset where we have full control on spurious correlations we desire to keep private from the model, providing a theoretical analysis and observing DisP's effectiveness compared to the theory. Then, in Sec.~\ref{sec:exper} we test the effectiveness of DisP on real datasets and finally Sec.~\ref{sec:conclusion} draws the conclusions.

%% file: sections/2_method.tex
\section{Methods and procedures}
\label{sec:method}

In this section, after introducing the notation, we present DisP, our proposed regularization term, whose aim is to regularize the deep features aiming at reducing the propagation of the chosen private features onward in the deep model. Towards this end, we are going to support the description of our proposed technique by introducing a toy dataset, biased-MNIST, where we have full control on the impact of some private features over the learning/inference process. Despite such dataset has been originally designed for debiasing applications, we can consider the biased features as the private ones. Then, we will introduce our DisP term and we will observe its impact during learning.

\subsection{Preliminaries}
\begin{figure}
    \centering
    \includegraphics[width=\columnwidth]{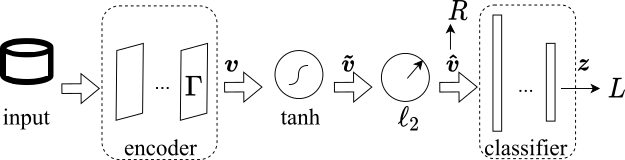}
    \caption{Representation for a generic trained model. The features for DisP are extracted at the output of the encoder ($\Gamma$ layer), and then tanh activated and normalized to have unitary norm.}
    \label{fig:structure}
\end{figure}
\begin{table}
   \center
    \caption{Overview on the notation used in this work.}
    \begin{tabular}{c l}
        \toprule
        \textbf{Symbol} & \textbf{Meaning}\\
        \midrule
        $M$                     & minibatch size\\
        $M_{t, p}$              & cardinality of the samples having the same target $t$\\
                                &and the same private feature $p$\\
        $C$                     & cardinality of targets\\
        $N_{\Gamma}$            & output size of $\Gamma$\\
        $\boldsymbol{v}_i$      & $i$-th sample in the minibatch, after $\Gamma$\\
        $\boldsymbol{\hat{v}}_i$      & $i$-th sample in the minibatch, after the normalization layer\\
        $\mathcal{T}(\boldsymbol{a}_i)$      & function which extracts the target class of $\boldsymbol{a}_i$\\
        $\mathcal{P}(\boldsymbol{a}_i)$ & function which extracts the private feature class of $\boldsymbol{a}_i$\\
        $\delta_{ab}$           &Kronecker delta\\
        $\bar{\delta}_{ab}$      &$1 - \delta_{ab}$\\
        $\langle \boldsymbol{a}, \boldsymbol{b} \rangle$   &scalar product between $ \boldsymbol{a}$ and $ \boldsymbol{b}$ transposed.\\
        $\mathbb{P}(a)$                  & probability of $a$\\
        $\mathbb{H}(a)$                  & entropy of $a$\\
        $\mathbb{I}(a,b)$                & mutual information between $a$ and $b$\\
        
        \bottomrule
    \end{tabular}
    \label{tab:notationover}
\end{table}
Let us assume we focus our attention to the output of $\Gamma$, which is the output layer of the encoder for our model (Fig.~\ref{fig:structure}). We say the output of $\Gamma$ for the $i$-th sample is $\boldsymbol{v}_i \in \mathbb{R}^{N_{\Gamma}\times M}$. DisP will be a regularization term, computed over features extracted from $\Gamma$ and undergoing proper activation (tanh and $\ell_2$ normalization, as displayed in Fig.~\ref{fig:structure}), to be minimized besides the overall loss of the model, computed at the output of the classifier. The effect we desire to achieve is to have the part of the deep model before $\Gamma$ to select features which are important to solve the target task, filtering all of those containing the private information. The overall objective function we aim to minimize with our learning is the classical formulation
\begin{equation}
    \label{eq:uprule}
    J = \eta L + \gamma R_{\perp},
\end{equation}
where $L$ is the loss function for the trained task, $\eta$ and $\gamma$ are positive hyperparameters and $R_{\perp}$ is our proposed DisP. A summary for the notation used through this work is provided in Table~\ref{tab:notationover}: as a standard appoach, bold values mark vector quantities.

\subsection{Bottleneck layer}
\label{sec:botlayer}
In order to address our regularization term, we need first to decide the layer where we wish to apply our DisP regularization, whose task is to prevent information leakage (or in other words, what $\Gamma$ should be). Due to the complexity of currently-deployed deep neural network architectures with residual layers~\cite{He2016DeepRL, Huang2017DenselyCC} and skip connections~\cite{Ronneberger2015UNetCN}, and consider some theoretical frameworks analyzing the information flow in a deep learning model~\cite{Tishby2015DeepLA, Goldfeld2020TheIB}, a reasonable choice for our elected information bottleneck layer is the one before the classification layers, or in other words, between the encoder and the classifier.\\
For our purposes, we desire the extracted features $\boldsymbol{v}$ to have unit norm; hence, we insert in the model a \emph{normalization layer}:
\begin{itemize}
    \item First, it takes the pre-activation $\boldsymbol{v} \in \mathbb{R}^{N_{\Gamma}\times M}$ and a tanh activation function is applied to it:
    \begin{equation}
        \tilde{v}_{ij} = \tanh(v_{ij})
    \end{equation}
    After this, $\tilde{v}_{ij}\in (-1; +1)~\forall i,j$. 
    \item Then, an $\ell_2$ normalization is applied to the entire feature vector $\boldsymbol{\tilde{v}}_i$:
    \begin{equation}
        \boldsymbol{\hat{v}}_i  = \frac{\boldsymbol{\tilde{v}}_{i}}{\|\boldsymbol{\tilde{v}}_i\|_2}.
    \end{equation}
\end{itemize}
The DisP regularization function will be applied to the normalized $\boldsymbol{\hat{v}}_i$ values. The typical ReLU activation function allows feature values in the range $[0;+\infty)$. For our purposes, we need the feature values where DisP is applied to be limited: we will empirically observe that including this activation at the bottleneck layer does not significantly impact over the performance of the model.

\subsection{Disentangling the Private information}
\label{sec:regu}

\begin{figure}
	\centering
    \begin{minipage}{0.45\columnwidth}
        \centering
        \includegraphics[trim= 0 0 130 0, clip,width=\textwidth]{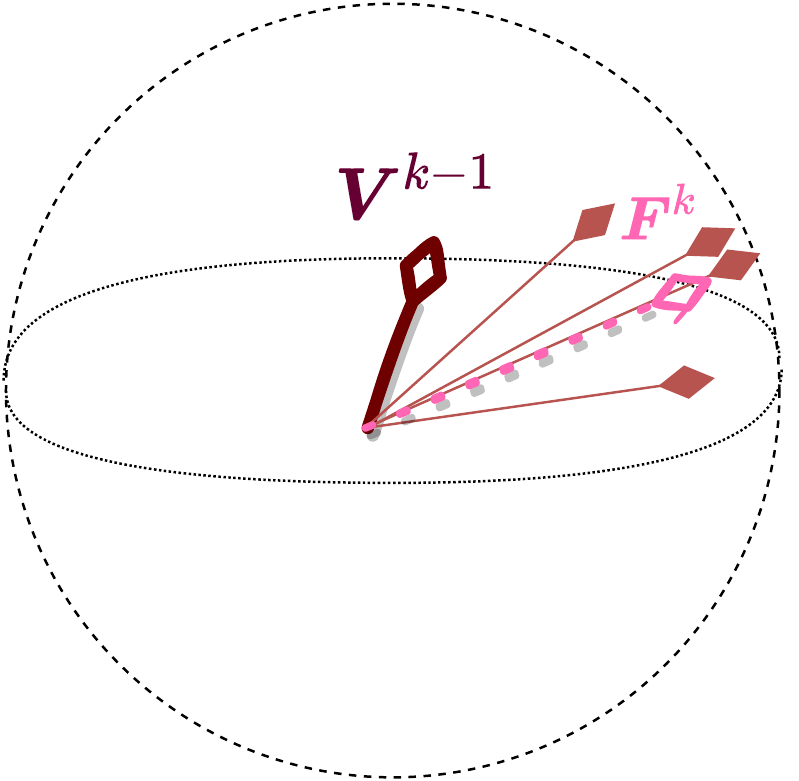}
        \caption*{(a)}
    \end{minipage}
    \begin{minipage}{0.45\columnwidth}
        \centering
        \includegraphics[width=\textwidth]{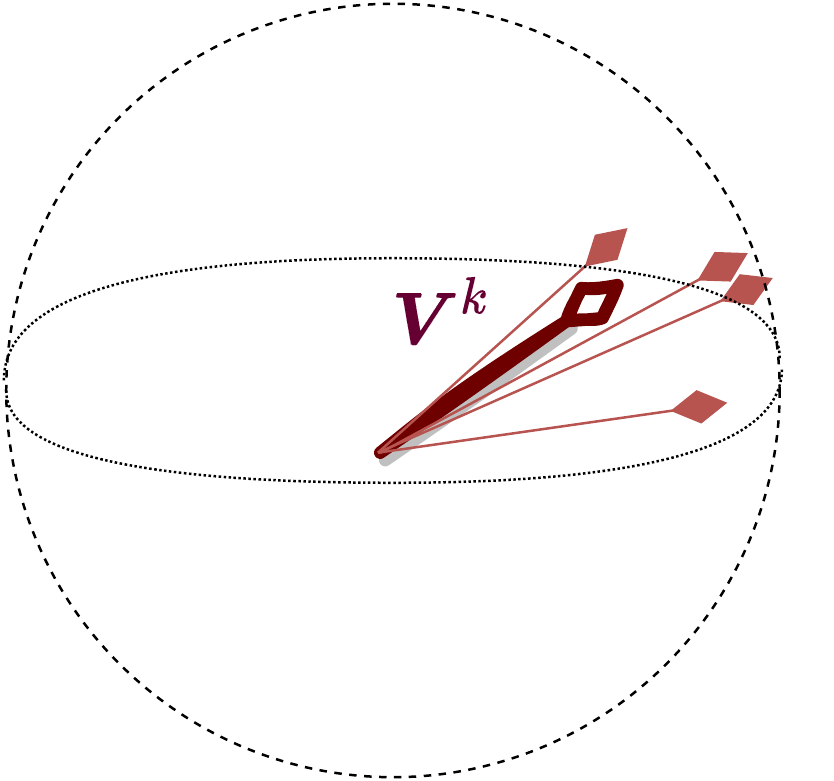} 
        \caption*{(b)}
    \end{minipage}
    \begin{minipage}{0.45\columnwidth}
        \centering
        \includegraphics[width=\textwidth]{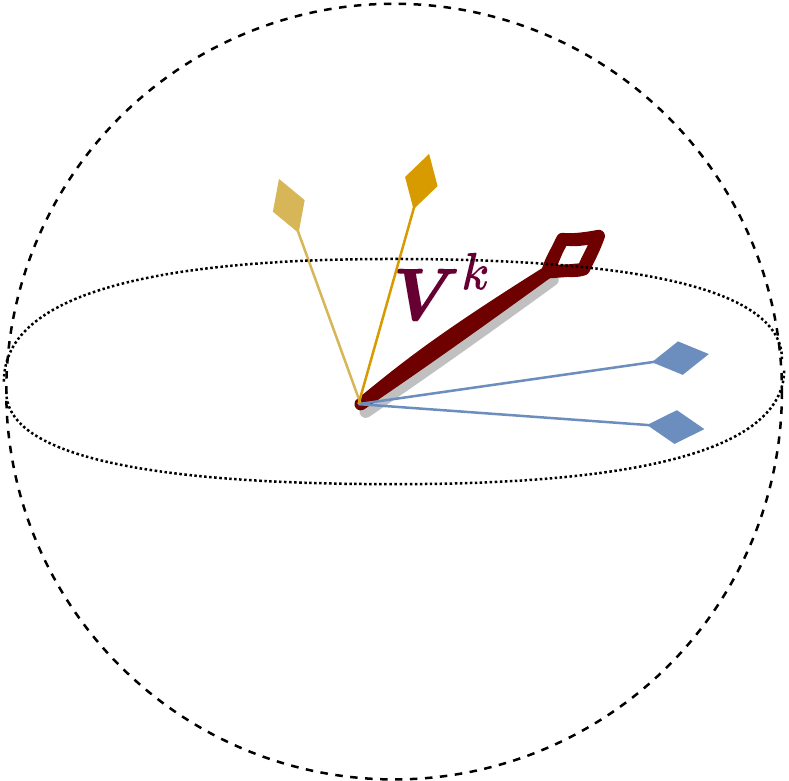}
        \caption*{(c)}
    \end{minipage}
    \begin{minipage}{0.45\columnwidth}
        \centering
        \includegraphics[width=\textwidth]{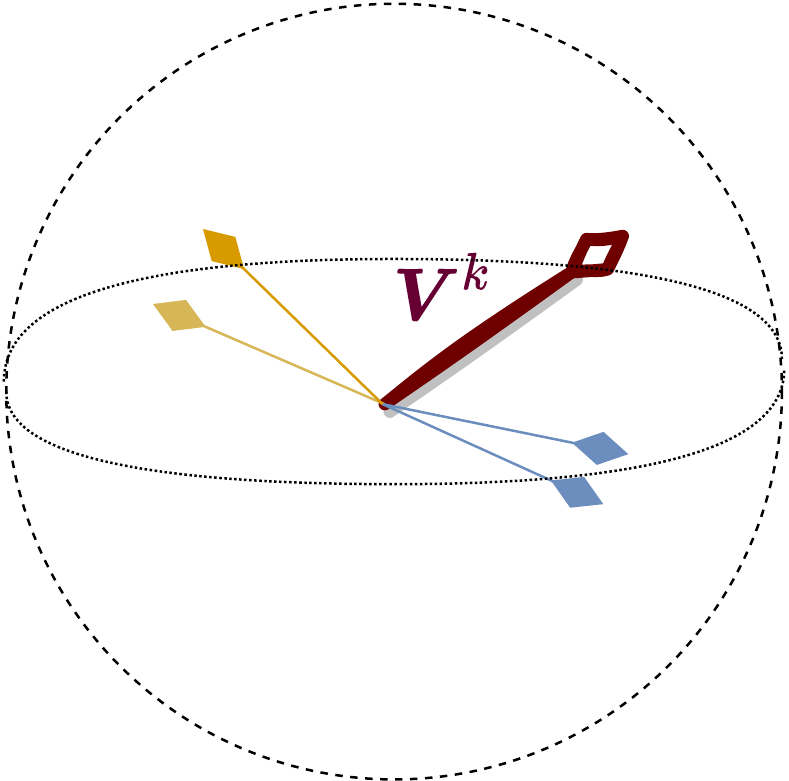} 
        \caption*{(d)}
    \end{minipage}
    \caption{Effect of the $R_{\perp}^{mem}$ term. Same arrow's point represents same private class $p$ while the same arrow's color indicates same target $t$. First the average $\boldsymbol{F}_{t,p}^{k}$ for the current mini-batch is computed (a), then the memory term $\boldsymbol{V}_{t,p}^{k}$ is provided (b). On the features with same bias but different target (c) the penalty $R_{\perp}^{mem}$ is applied (d).}
    \label{fig:memReffect}
\end{figure}

In order to disentangle the private features at training time, we are required two terms:
\begin{itemize}
    \item the \emph{average} features for every target $t$ and every private class $p$, over all the dataset;
    \item the \emph{current} features for every target $t$ and every private class $p$, over the computed minibatch.
\end{itemize}
What we aim to accomplish here is to disentangle features belonging to the same private class $b$ but different target classes. In this way, differently from other works like \cite{tartaglione2021end} and \cite{Tartaglione2020ANA}, we will minimize the risk of destroying non-private information in the process, at the same time preventing private data leakage and maintaining the performance high.\\
For this reason, we first update the average private features by computing the average features for the current mini-batch:
\begin{equation}
    \label{eq:F}
    \boldsymbol{F}_{t,p}^{k} = \frac{1}{M_{t,p}}\sum_{i} \boldsymbol{\hat{v}}_i \cdot \delta_{t,\mathcal{T}(\boldsymbol{\hat{v}}_i)} \cdot \delta_{p,\mathcal{P}(\boldsymbol{\hat{v}}_i)}
\end{equation}
with $k$ as a mini-batch index, and we update the memory term
\begin{equation}
    \label{eq:V}
    \boldsymbol{V}_{t,p}^{k} = (1-\beta)\cdot \boldsymbol{V}_{t,p}^{k-1} + \beta \cdot \boldsymbol{F}_{t,p}^{k},
\end{equation}
where $\beta$ is the momentum coefficient. Then, we can compute a memory-based disentangling term for the current mini-batch
\begin{equation}
    \label{eq::regumem}
    R_{\perp}^{mem} = \frac{1}{M\cdot (T-1)}\sum_{t=1}^T \sum_{i=1}^M \left| \langle \boldsymbol{\hat{v}}_i, \boldsymbol{V}_{t,p}^{k}\rangle \right|\cdot \bar{\delta}_{t,\mathcal{T}(\boldsymbol{\hat{v}}_i)} .
\end{equation}
\begin{figure}
	\centering
    \begin{minipage}{0.45\columnwidth}
        \centering
        \includegraphics[width=\textwidth]{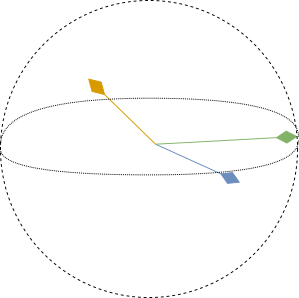}
        \caption*{(a)}
    \end{minipage}
    \begin{minipage}{0.45\columnwidth}
        \centering
        \includegraphics[width=\textwidth]{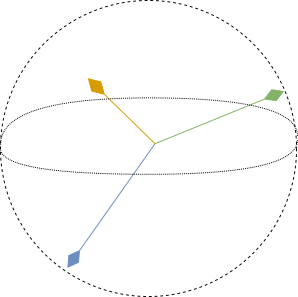} 
        \caption*{(b)}
    \end{minipage}
    \caption{Effect of the $R_{\perp}^{batch}$ term. For each $i$-th example, the others having same bias but different target are selected - in (a), the yellow vector is out $i$-th. Then, these vectors are each-other de-correlated (b).}
    \label{fig:batchReffect}
\end{figure}
Minimizing \eqref{eq::regumem} equals to minimizing the average of the correlations between features belonging to the same private class but different target classes. A visual representation on the computation of this term is depicted in Fig.~\ref{fig:memReffect}. However, minimizing this term is not a sufficient condition to de-correlate all the features: there can still be formed clusters mapping the private information. For example, looking at Fig.~\ref{fig:memReffect}d, the features in blue and yellow are de-correlated from $\boldsymbol{V}_{t,p}^{k}$, but the two classes are still anti-correlated. Towards this end, we also have a term which de-correlates \emph{all} the features within the same mini-batch, mapped to the same bias but having different targets:
\begin{equation}
    \label{eq::regubatch}
    R_{\perp}^{batch} = \frac{1}{M} \sum_{i=1}^M \frac{1}{\sum\limits_{t \neq \mathcal{T}(\boldsymbol{v}_i)} M_{t, \mathcal{P}(\boldsymbol{\hat{v}}_i)}}
    \cdot\sum_{j} \left|\left \langle \boldsymbol{\hat{v}}_i, \boldsymbol{\hat{v}}_j \right\rangle\right| \cdot \bar{\delta}_{t,\mathcal{T}(\boldsymbol{\hat{v}}_i)}
\end{equation}
This term de-correlates all the features belonging to the same private class but different target classes within the mini-batch. A visual effect of $R_{\perp}^{batch}$ is shown in Fig.~\ref{fig:batchReffect}. Having this term alone, with no memory term, for a large dataset, might result in an oscillatory state, where some features might be re-correlated after a certain number of epochs. Towards this end, the memory term in \eqref{eq::regumem}, despite modeling only the average correlation among all the private features, helps in a long-range average de-correlation. The two contributions in \eqref{eq::regumem} and \eqref{eq::regubatch} are simultaneously minimized, and together they constitute the DisP regularization term:
\begin{equation}
    \label{eq::regu}
    R_{\perp} = \gamma^{mem}\cdot R_{\perp}^{mem} + \gamma^{batch} \cdot R_{\perp}^{batch}
\end{equation}
where $\gamma^{mem}$ and $\gamma^{batch}$ are two positive hyper-parameters, which have typically the same value (in such cases, we will refer to those simply as $\gamma$). 

%% file: sections/3_attacks.tex
\section{Attacks}
\label{sec:attacks}
\begin{figure}
	\centering
    \begin{minipage}{\columnwidth}
        \centering
        \includegraphics[trim={100 0 0 0}, width=0.9\textwidth]{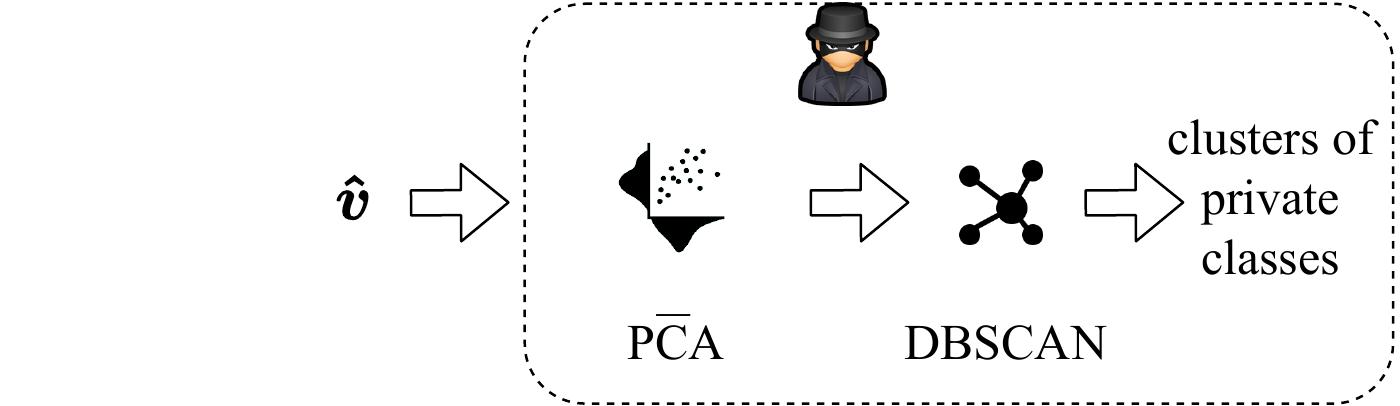}
        \caption*{(a)}
    \end{minipage}
    \begin{minipage}{\columnwidth}
        \centering
        \includegraphics[trim={100 0 120 0}, width=0.7\textwidth]{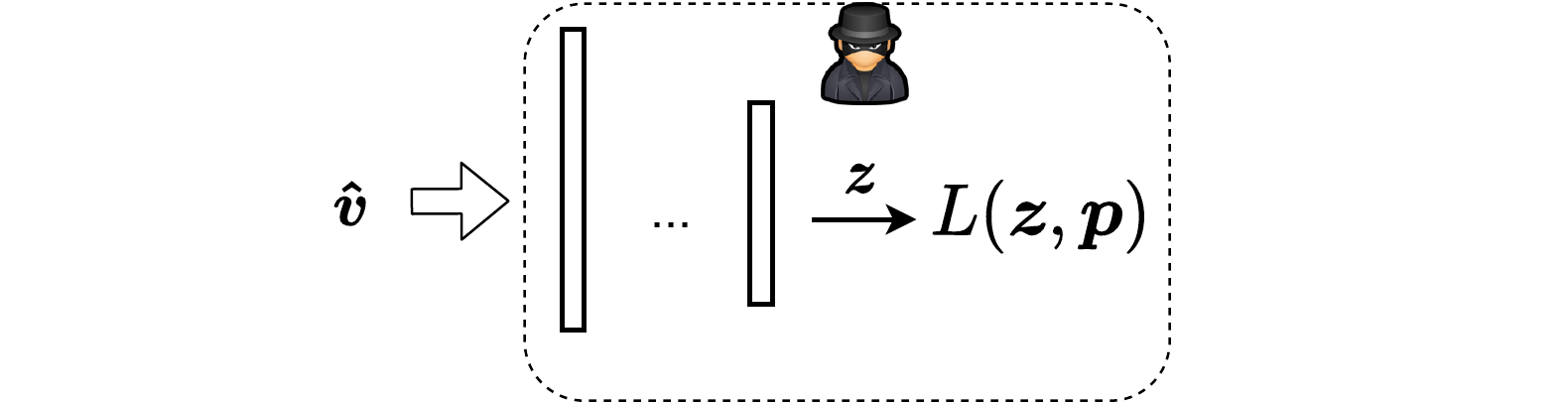} 
        \caption*{(b)}
    \end{minipage}
    \caption{Possible class inference attacks on the regularized $\boldsymbol{\hat{v}}$: unsupervised attack (a) and worst-case supervised attack (b).}
    \label{fig:attacks}
\end{figure}
In order to assess potential attacks aiming at retrieving the private features, we assume a potential attacker has access to $\boldsymbol{\hat{v}}_i$.\\
Here, a typical approach to mine the private information is to use an unsupervised approach. Towards this end, a possible attack involves the following steps:
\begin{itemize}
    \item Compute the PCA over the set of the available data. Through this, it is possible to reduce the dimensionality of the data. A typical good choice is to extract the components such that the energy content is at least the $95\%$ over the total energy. We assume a worst-case scenario where the attacker has \emph{all} the possibly available data (hence, the whole of the training/test sets);
    \item Run DBSCAN to find data dependencies. Despite an attacker might know $\boldsymbol{\hat{v}}_i$ is normalized over a unitary hyper-sphere and might be tempted in using special clustering algorithms designed for this scenario like spherical k-means~\cite{buchta2012spherical}, the typical high-dimensionality of the data makes more useful to apply first a PCA to reduce such dimensionality. After the PCA is applied, the spherical constraint typically does not hold anymore; hence, a more general approach, like DBSCAN, which is able to capture non-trivial data dependencies, should be deployed.
\end{itemize}
Once clusters of data are found, we can look at the content of each of these clusters: if all of these contain an homogeneous number of datapoints belonging to the same private class, then the private information has not been extracted, and the attacker failed. On the contrary, if there are clusters containing a high percentage of datapoints with the same private class, then the attack has success and the private information has been retrieved. We call this \emph{unsupervised attack}.\\
As an upper-bound case, let us assume the attacker has the training set data provided with the private class label, which is hence known. In such a case, the attacker is willing to train a classifier from $\tilde{y}_i$, attempting to learn the relationship between the input $\tilde{y}_i$ and the private features $b$. Such an attack has success if the trained model, which in our case will be a multi-layer preceptron (MLP), is able to generalize well on the test set data. If the information about the private classes has been erased, the MLP will be forced in a \emph{memorization} state, where the training set will necessarily overfit over the training set, if the number of parameters in it is sufficiently large. We call this \emph{supervised attack} and constitutes an upper bound to the learnable private information given our testing methods. A graphical summary of the attacks is provided in Fig.~\ref{fig:attacks}.

%% file: sections/4_biasedmnist.tex
\section{A controlled experiment}
\label{sec:control}
In this section we are going to conduct experiments over biased-MNIST, a dataset where we have direct access and control to parameters directly tuning correlations between input features and some extra features not directly correlated to the target task we desire to train a model on. This allows us to address a simplified theoretical framework which gives us insights over the effect of DisP over the information extracted from the available data.

\subsection{The biased-MNIST dataset}
\label{sec:biasedmnist}
\begin{figure}
    \centering
    \includegraphics[width=\columnwidth]{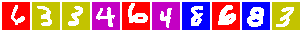}
    \caption{Some samples from biased-MNIST~\cite{bahng2019rebias}: the background colors (our private feature) highly-correlate with the digit classes (the target).}
    \label{fig:biased-mnist}
\end{figure}
This dataset has been recently proposed by Bahng~\emph{et~al.}~\cite{bahng2019rebias} for debiasing purposes; however, we can consider the biased features of such a dataset as the features we desire to hide from the classifier. This dataset is constructed from MNIST~\cite{lecun2010mnist} by injecting a color into the images background, as shown in Fig.~\ref{fig:biased-mnist}. In this dataset, we have ten different targets (hence, $C=10$). Each digit is associated with one of ten pre-defined colors, which will be our private feature. To assign the background color to an image of a given target class, the pre-defined color is selected with a probability $\rho$, and any other color is chosen with a uniform probability $(1-\rho)$:
\begin{equation}
    \begin{cases}
        \mathbb{P}(\mathcal{T}\left(\boldsymbol{x}) = i | \mathcal{P}(\boldsymbol{x}) = i\right ) = \rho \\
        \mathbb{P}(\mathcal{T}\left(\boldsymbol{x}) \neq i | \mathcal{P}(\boldsymbol{x}) = i\right ) = \frac{1}{9} (1 - \rho).
    \end{cases}
\end{equation}
Straightforwardly we can assume
\begin{align}
 \mathbb{H}(T) &= \mathbb{P}(t) \log_{10} (0.1) = 1 \\
 \mathbb{H}(P) &= \mathbb{P}(p) \log_{10} (0.1) = 1,
\end{align}
where $T$ and $P$ are the random variables associated to the target and the private class, respectively. From this, can write the conditional entropy
\begin{equation}
    \mathbb{H}(P|T) = \rho \cdot \log_{10}\left[\frac{1-\rho}{9 \cdot\rho}\right] + \log_{10}\left[\frac{9}{1-\rho}\right]. 
\end{equation}
Here, we can write the mutual information
\begin{equation}
    \mathbb{I}(P, T) = 1 - \rho \cdot \log_{10}\left[\frac{1-\rho}{9 \cdot\rho}\right] + \log_{10}\left[\frac{9}{1-\rho}\right].
\end{equation}
\begin{figure}
    \centering
    \includegraphics[width=0.8\columnwidth]{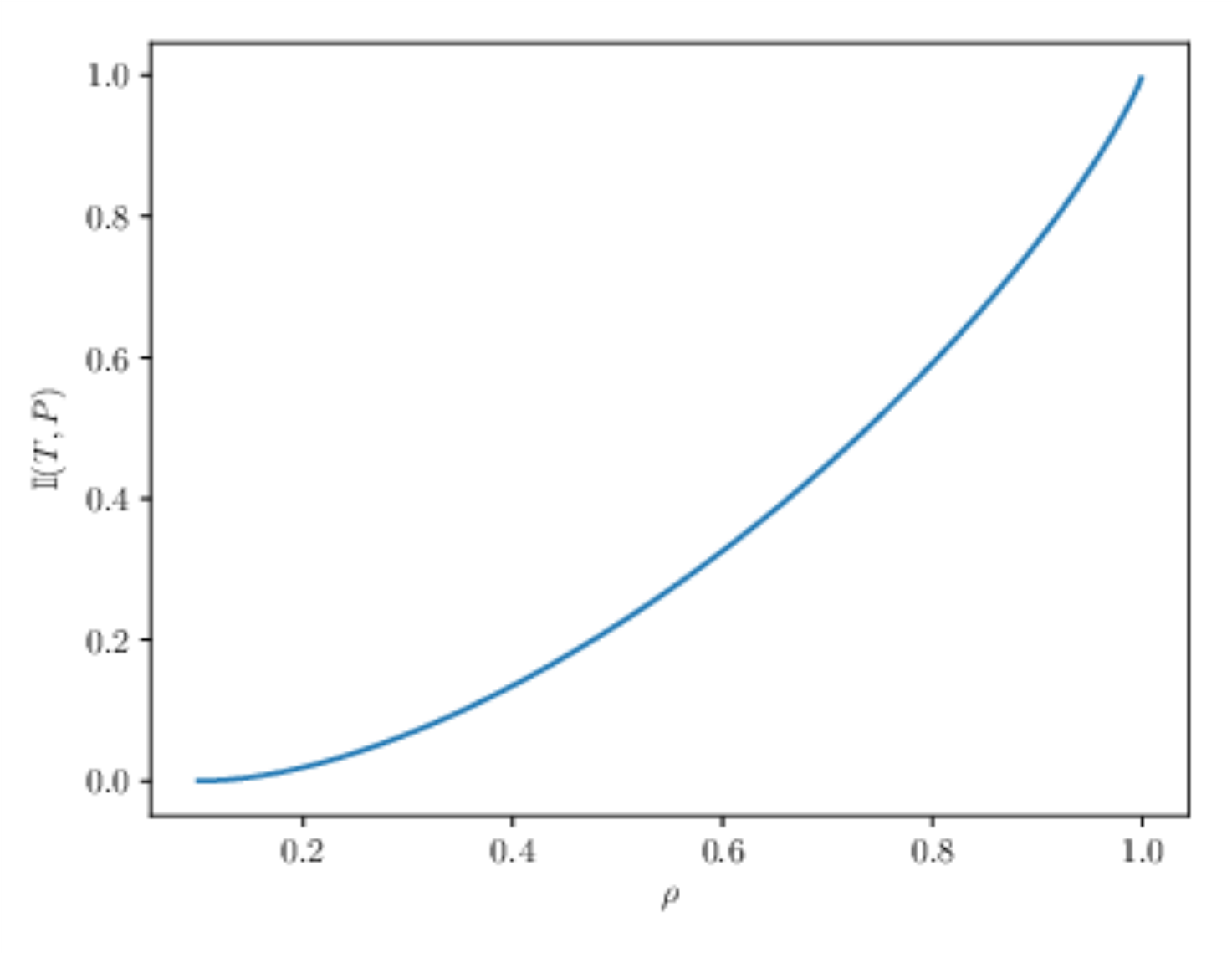}
    \caption{Plot of the mutual information between $T$ and $P$.}
    \label{fig:theoryMIXY}
\end{figure}
Figure~\ref{fig:theoryMIXY} shows the mutual information between the target X and the private feature Y. Of course, the more the target correlates to the private feature, the highest the chances of having some information leakage. In the next we are going to quantitatively analyze the information leakage on a trained model, introducing the random variable Z which models the output of the model.

\subsection{Expected information leakage in biased-MNIST}
\label{sec:ileak}
\begin{figure}
    \centering
    \includegraphics[width=\columnwidth]{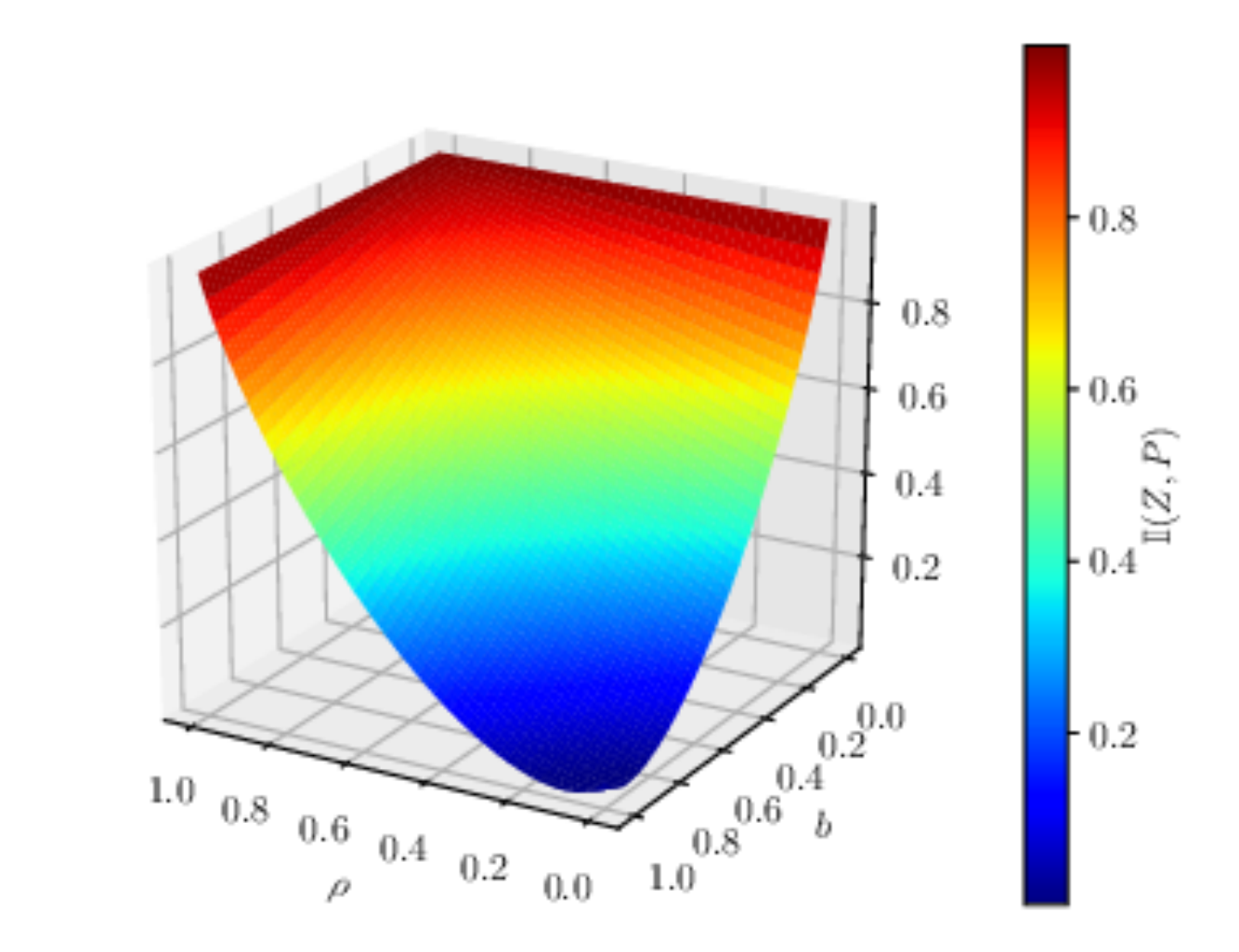}
    \caption{Plot of the mutual information between the private feature and the output of the model. In this case all the $b$ are equal for visualization purposes.}
    \label{fig:theoryMIYZ}
\end{figure}

Let us assume here a real case where the trained model is not a perfect learner, meaning that
\begin{equation}
    \mathbb{H}(Z|T) \neq 0
\end{equation}
where $Z$ is the random variable associated to the output $\boldsymbol{z}$. The model does not correctly classify the target for two reasons.
\begin{itemize}
    \item It gets confused by the private features, and it tends to learn to classify on top of those. We model the tendency of learning private features as $b$: the lower this parameter, the most the model learns the features we desire to keep private, introducing error in the model. 
    \item Some extra error, un-related to the bias features, which can be due to stochastic unbiased effects, or to underfit, or to other high-order dependencies between data. We are not strictly interested to model such an effect for our purposes, and we consider this extra error always be zero.
\end{itemize}
We can now write the joint probability:
\begin{align}
   \mathbb{P}(t, p, z) =& \frac{1}{10}\cdot \left[\delta_{tpz} \rho + (1 - \delta_{tz})\delta_{tp} \frac{(1-b)(1-\rho)}{9}+\right.\nonumber\\ 
   &\left .+(1-\delta_{tp})\delta_{tz}\frac{b(1-\rho)}{9} \right]
\end{align}
where $b\in[0; 1]$. From this, we can compute the marginal probability
\begin{align}
   \mathbb{P}(p, z) = &\frac{1}{10}\cdot \left\{\delta_{pz} \left[\rho + \frac{1-\rho}{9}\sum_{m}b(1 - \delta_{pm})\right] +\right .\nonumber \\
   &\left. +(1-\delta_{pz}) \left[ (1 - b) \frac{1 - \rho}{9} \right] \right\}.
\end{align}
From this, we can plot the mutual information between the neural network prediction and the source (Figure~\ref{fig:theoryMIYZ}). \\
We observe that for high $\rho$ values the mutual information between the information the model learns and the private classes is high, independently from how much the model learns to extract from the private features ($b$). Our objective here is to discourage the leakage of private information on new data: if we keep the value of $b\approx 0.9$, the model still is able to keep good performance, but when tested on un-correlated data ($\rho=0.1$), the mutual information drops. On the contrary, for low values of $b$, $\mathbb{I}(Z,P)$ remains high, meaning that there is private information leakage.

\subsection{Training with DisP}
\label{sec:bmnistperf}
We use the network architecture proposed by Bahng~\emph{et~al.}~\cite{bahng2019rebias}, consisting of four convolutional layers with $7\times 7$ kernels. As a bottleneck layer here we choose the average pooling layer, before the fully connected classifier of the network. We optimize the model using stochastic gradient descent with learning rate 0.1, batch size 100 and weight decay 1e-4, for 50 epochs. 
\begin{table}
    \caption{Experiments on biased-MNIST. (*) values rounded to the closest decimal.}
    \label{tab:measuresbMNIST}
    \small
	\renewcommand{\arraystretch}{1.2}
	\centering
		\begin{tabular}{c c c c c c c c c c c c c c}
        \toprule
        \multirow{2}{*}{$\rho$} & \multirow{2}{*}{$\gamma$}  & \multirow{2}{*}{$b$}   & \multirow{2}{*}{$R_{\perp}$}     &Accuracy $[\%]$\\
                &           &               &               &$(\rho=0.1)$\\
        \midrule
        \multirow{3}{*}{0.1}
            &0            &0.9*     & 0.01  &98.34\\
            &0.1          &0.9*  & 1e-3 &98.21\\
            &0.2          &0.9*  & 3e-5 &98.03\\
        \midrule
        \multirow{3}{*}{0.99}
            &0           & 0.21    & 0.44 &89.10\\
            &0.1         & 0.75    & 0.12 &95.01\\
            &0.2         & 0.83    & 0.07 &93.73\\
        \bottomrule
	\end{tabular}
\end{table}
Table~\ref{tab:measuresbMNIST} shows the experimental results achieved training 5 models per configuration. In this case, we are also able to provide an empirical estimation on $b$ (which is the tendency of the model to learn the private features, as discussed in Sec.~\ref{sec:ileak}).\\
In the case where there is no correlation between target and private feature ($\rho=0.1$), and the tendency $b$ is low, also the information relative to the private feature correlation (R) is typically low, and further minimizing it is not harming or even improving the generalization capability of the model.\\
However, when the correlation between target and private feature is extremely high ($\rho=0.99$), the tendency value $b$ drops, meaning that the model is learning the private information in place of the target. This is due to the very high correlation between target and private feature, which is $\rho$ by definition. In such a case, including the regularization term in the cost function is minimizing the correlation term R and is also beneficially impacting the accuracy of the model itself. Indeed, the tendency $b$ is increasing, meaning that the model is no longer learning the private feature but is looking at the correct features to extract. 

\subsection{Attacks}
\begin{figure}
    \centering
    \includegraphics[width=\columnwidth]{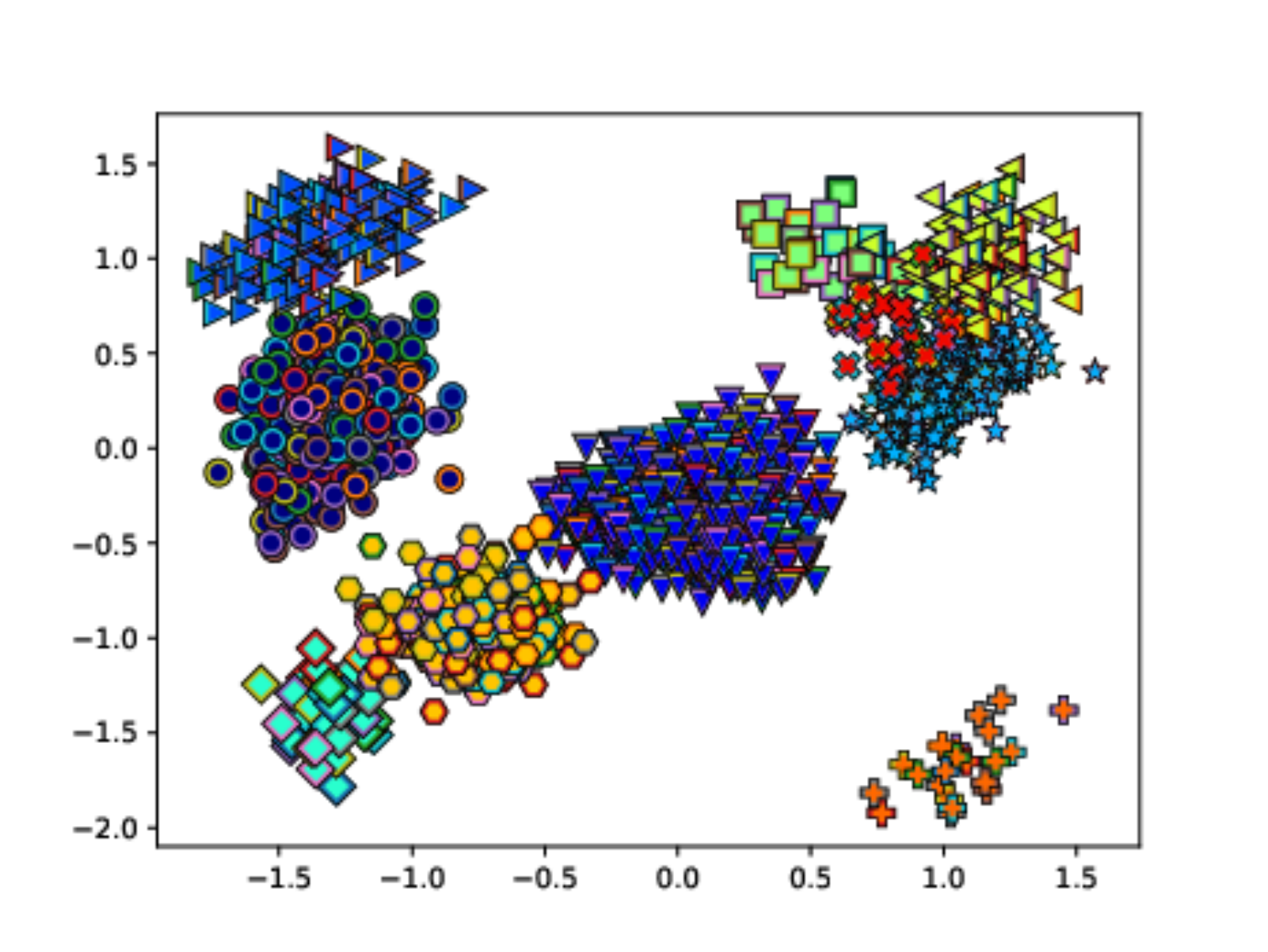}
    \caption{PCA+DBSCAN output for a model regularized with DisP and trained on biased-MNIST, with $\rho=0.99$ and $\gamma=0.1$, on the test set. Here the first two principal components are displayed. A different symbol is a different cluster found by DBSCAN, internal color is the target class while edge color is the private class.}
    \label{fig:uMNIST}
\end{figure}
Considering that in this case we have a theoretical framework estimating the upper-bound over the private feature which can be extracted, we will here conduct the unsupervised attack only. We run attacks here for $\rho=0.99$, with $\gamma=0$ (which is our baseline model) and $\gamma=0.1$. We run the attacks over the models showing the median accuracy over the test set. Figure~\ref{fig:uMNIST} qualitatively shows the outcome of the clustering algorithm. In particular, the different shape represents the cluster found by the PCA+DBSCAN approach, the internal color is the target while the border color is the private features. Even qualitatively, we observe that without applying DisP, there is a significant part of the clusters which map to the private feature, while applying DisP this percentage drops dramatically, averagely to the $14\%$.

\subsection{Ablation study}
\begin{table}
    \caption{Ablation study on biased-MNIST ($\rho=0.99$).}
    \label{tab:ablation}
    \small
	\renewcommand{\arraystretch}{1.2}
	\centering
		\begin{tabular}{c c c c c c c c c c c c c c}
        \toprule
        $\gamma^{mem}$ & $\gamma^{batch}$    & $R_{\perp}^{mem}$ & $R_{\perp}^{batch}$    &$R_{\perp}$&Accuracy $[\%]$\\
                       &                     &                    &                       &          &$(\rho=0.1)$\\
        \midrule
        0.1& 0   &0.04  &   0.18  & 0.22    & 91.23      \\
        0  &0.1  &0.07  & 0.09  & 0.16      & 92.54      \\
        \bottomrule
	\end{tabular}
\end{table}
Here we propose an ablation study on DisP. In particular, we evaluate the impact of the two regularization contributes, the memory term $R_{\perp}^{mem}$ \eqref{eq::regumem}, and the mini-batch term $R_{\perp}^{batch}$ \eqref{eq::regubatch}. We evaluate this in the challenging scenario where $\rho=0.99$. Table~\ref{tab:ablation} shows the ablation study results obtained: even though in some cases the weight in the regularization loss ($\gamma$) is zero, we still measure and report it. While applying singularly the two contributions is indeed minimizing the overall $R$ in both cases, the lowest value for $R$, and in this case, the highest accuracy, is achieved applying the same regularization contribution to both. Of the two terms, the largest contribution is overall provided by $R_{\perp}^{batch}$, which disentangles all the feature vectors in the name minibatch.

%% file: sections/5_experiments.tex
\section{Experiments on real datasets}
\label{sec:exper}
In this section we experiment DisP's effectiveness over two real datasets. Our goal here is to train the ANN model on the target classification task applying our regularization strategy over some features we wish to keep private (in our case, in order to maintain consistency over the experiments, we want to keep private the information over gender). We will show that, in our context, such an information is detectable in more trivial cases (like in face images datasets) and in less straightforward cases (like in chest X-ray radiographic images).\\
Our training and inference algorithms are implemented in Python, using PyTorch~1.12 and a RTX2080~Ti NVIDIA GPU with 11GB of memory has been used for training and inference.\footnote{The source code will be made available upon acceptance of the article.} All the used hyper-parameters have been tuned using grid search.\\
The results will be presented for two different datasets: Celeb-A and SIIM-FISABIO-RSNA COVID-19 (SFR). After training has been completed, the attacks proposed in Sec.~\ref{sec:attacks} will evidence the possibility of extracting the private information from the bottleneck features $\boldsymbol{\hat{y}}_i$. As a chosen architecture, we select the model which has achieved the lowest validation loss along the training.

\subsection{Datasets and setup}

\textbf{CelebA.} This dataset~\cite{liu2015faceattributes} has been designed for face-recognition tasks, providing 40 attributes for every image. The dataset contains a total of 202.6k images and, following the official train-validation split, we obtain 162.7k images for training set, 19.9k images as validation set and 19.9k images for testing our models. For our training purposes, we use a ResNet-18 model, choosing the adaptive pooling layer as bottleneck layer $\Gamma$. The training has been performed using SGD, with an initial learning rate of 0.1, decayed by a factor 10 after no improvement over the validation set loss has been detected for 10 consecutive epochs. The training stops when the learning rate drops below 1e-3. We use batch size 100 with momentum of 0.9 and weight decay of 1e-5. Images are here re-scaled to $224\times 224$ resolution.\\
We select here as classification target the recognition of the \emph{Eyeglasses}, \emph{BlondHair} and \emph{Heavymakeup} attributes, while the gender is the private class. While the eyeglasses ($\rho=0.62$) and blondehair ($\rho=0.58$) attributes are relatively easy to disentangle from the gender, disentangling the heavymakeup attribute is a more challenging choice ($\rho=0.82$), as there is a high-correlation between these attributes (most of the women in the dataset have a heavy makeup).\\

\textbf{SIIM-FISABIO-RSNA COVID-19.} The currently available dataset\footnote{\url{https://www.kaggle.com/c/siim-covid19-detection/data}} comprises more than 6k chest X-ray (CXR) scans in DICOM format, anonymized according to the current GDPR guidelines. For study purposes, however, the metadata associated to these scans comprises information about the gender, which will be used as our private class. We split the dataset in a training set comprising 5k scans, a validation set of 300 scans and a test set of 600 scans. We train a DenseNet-121 model to classify over the ``Negative for Pneumonia'' and ``Typical Appearance'' classes. As a bottleneck layer here we choose, similarly to what chosen for CelebA, the adaptive pooling layer. The training has been performed using SGD, with an initial learning rate of 0.1, decayed by a factor 10 after no improvement over the validation set loss has been detected for 5 consecutive epochs. The training stops when the learning rate drops below 1e-3. We use batch size 16 with momentum of 0.9 and weight decay of 1e-4. The scans are converted using the meta-information contained in the DICOM files, and re-scaled to $448\times 448$ resolution.

\subsection{Results} 
\begin{table*}
    \caption{Results on the trained models with original train-validation-test split.}
    \label{tab:totalresults}
    \small
	\renewcommand{\arraystretch}{1.2}
	\centering
		\begin{tabular}{c c c c c c c c c c c c c c}
        \toprule
                &       &\multicolumn{3}{c}{\bf Training}&\multicolumn{3}{c}{\bf Attacks}\\
        Dataset & Target & $\gamma$ & $R_{perp}$     &Accuracy  & Unsup. & Sup.(1H) & Sup.(2H)\\
                &       &           &       &  [\%]     & [\%]  &  [Train\%-Test\%]  &[Train\%-Test\%]\\
        \midrule
        \multirow{6}{*}{CelebA}
            &\multirow{3}{*}{Blondhair}
                &0           &0.38      &95.68  &73.1  &85.8-79.3   &91.2-84.8\\
                &&0.1        &0.04      &95.88  &57.3  &86.3-64.8   &93.0-58.4\\
                &&0.2        &0.02      &95.32  &55.5  &83.7-61.4   &90.3-55.8\\
            \cline{2-8}
            &\multirow{3}{*}{Eyeglasses}
                &0           &0.08      &99.65  &51.3  &81.3-52.3   &88.9-51.2  \\
                &&0.1        &2e-3      &99.60  &53.2  &83.5-53.5   &89.1-52.3\\
                &&0.2        &3e-4      &99.60  &52.6  &82.1-51.7   &88.3-51.9\\
            \cline{2-8}
            &\multirow{3}{*}{Heavymakeup}
                &0           &0.39      &90.30  &84.1  &94.6-94.5   &94.6-94.6  \\
                &&0.1        &0.13      &90.33  &83.3  &94.1-94.1   &94.5-94.5\\
                &&0.2        &0.03      &89.79  &75.1  &93.6-93.3   &93.7-93.0\\
        \midrule
        \multirow{3}{*}{SFR}
            &\multirow{3}{*}{Pneumonia/Typical}
                &0           &0.47      &78.12  &75.2  &83.2-82.4   &85.7-81.9\\
                &&0.1        &0.12      &78.43  &53.6  &67.9-58.4   &69.5-60.6 \\
                &&0.2        &0.08      &78.02  &52.9  &65.6-58.9   &73.7-57.8\\
        \bottomrule
	\end{tabular}
\end{table*}

\begin{figure*}
    \captionsetup[subfigure]{position=b}
    \subcaptionbox{~\label{fig::R_covid}}{\includegraphics[width=.32\linewidth]{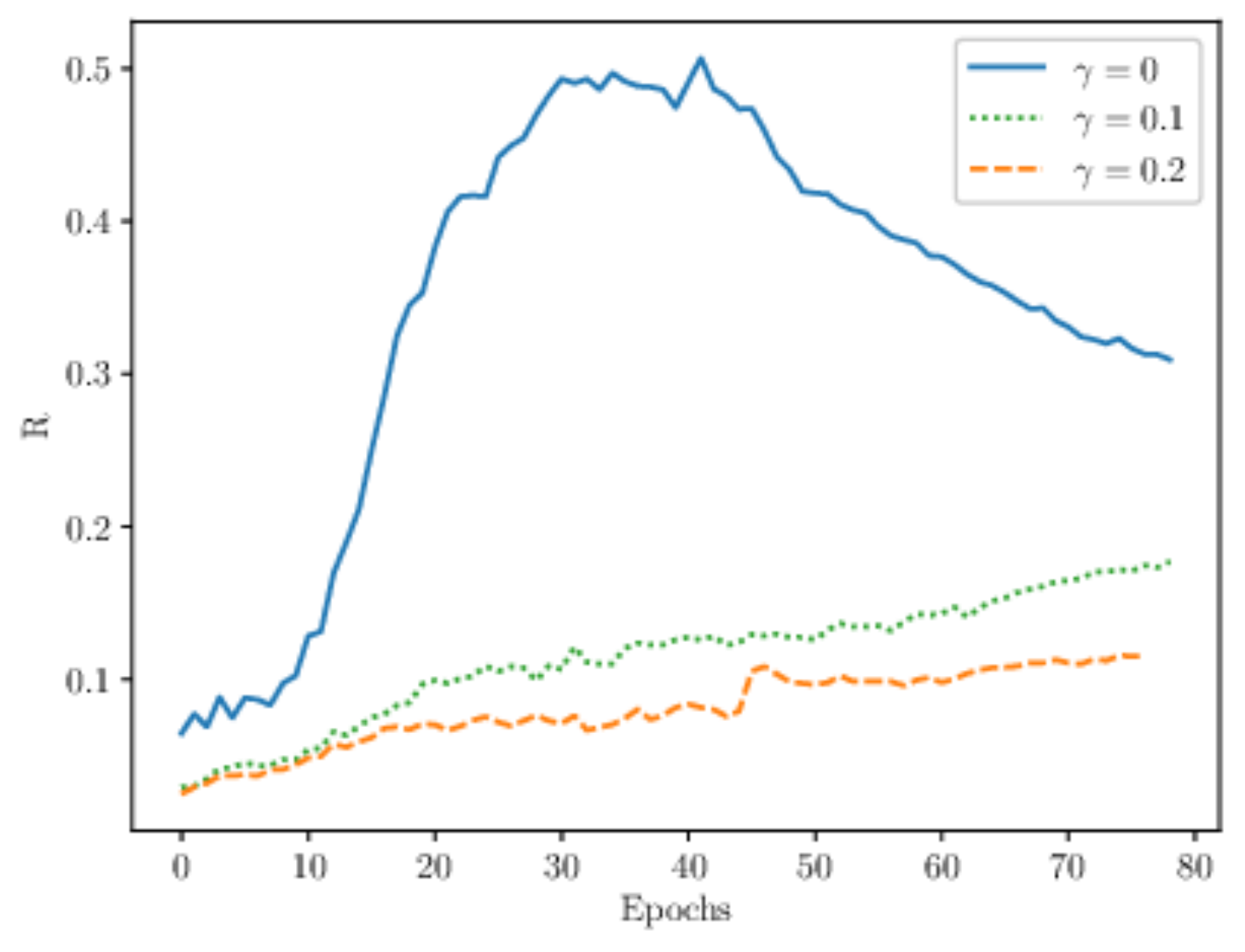}}
    \subcaptionbox{~\label{fig::validation_covid}}{\includegraphics[width=.32\linewidth]{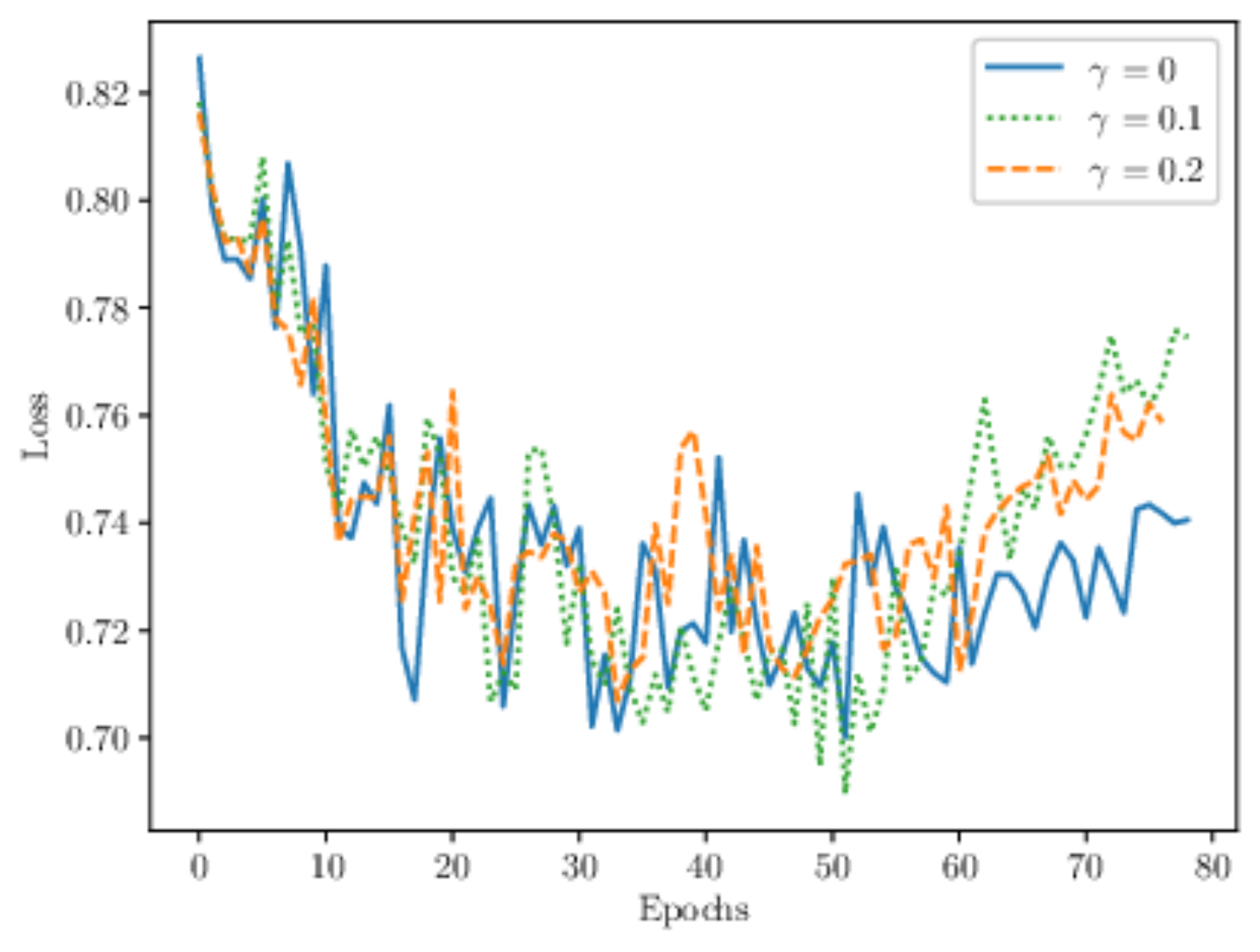}}
    \subcaptionbox{~\label{fig::test_covid}}{\includegraphics[width=.32\linewidth]{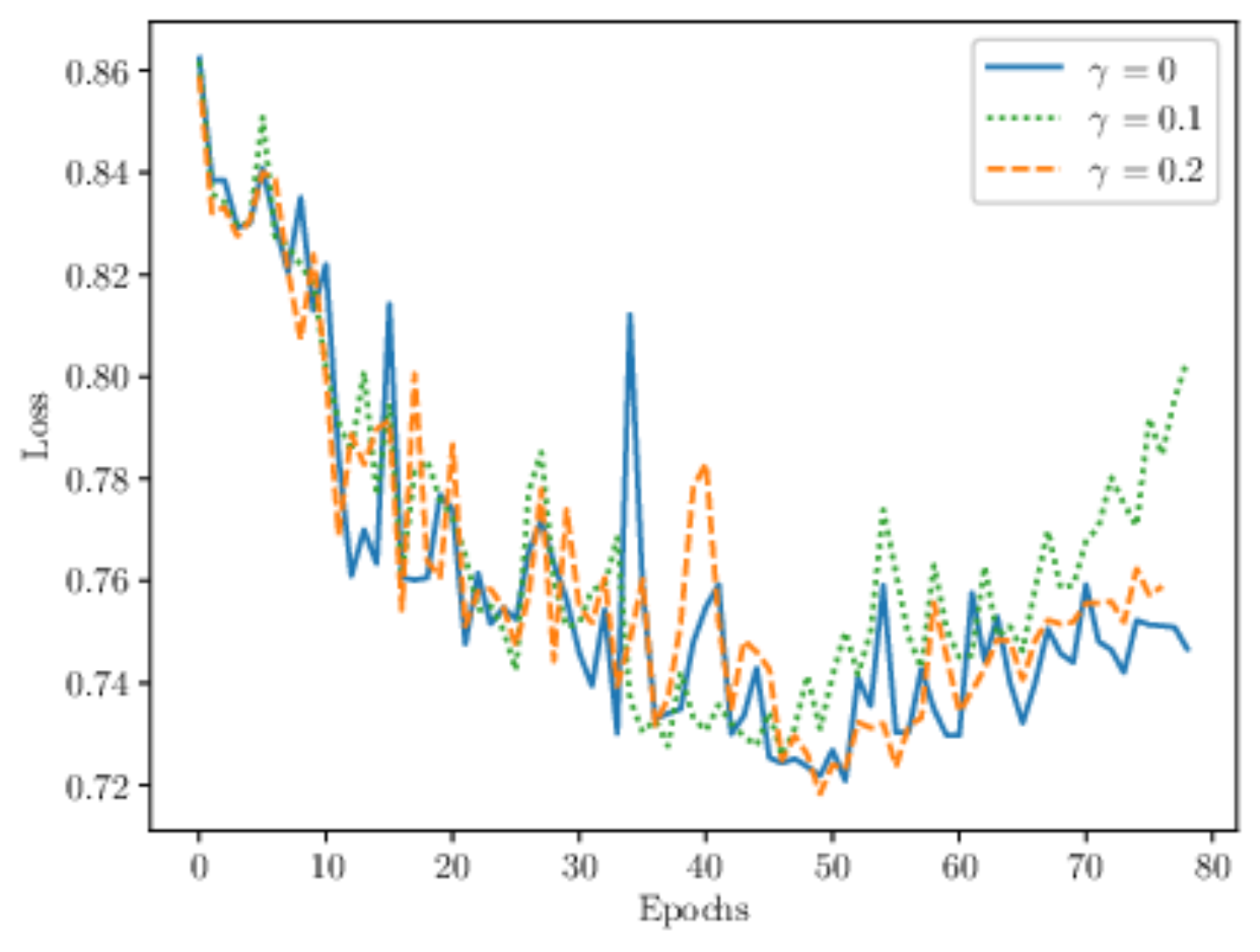}}
    \caption{Training on SFR: R value (a), validation set loss (b) and test set loss (c).}
    \label{fig::validationtrend}
\end{figure*}
Table~\ref{tab:totalresults} shows the achieved results. We report the performance on the training set, the accuracy in retrieving the private features with the unsupervised attack and the train/test accuracy when training a MLP model with one hidden layer of size 300 (1H) or two hidden layers of size 600-300 (2H). Looking at the pure performance at training time (or in other words, the performance for the given target) we observe that in most of cases, the use of DisP is not significantly harming the performance. Differently from what observed in biased-MNIST in Sec.~\ref{sec:bmnistperf}, there is overall no improvement in the performance when DisP is applied either. This means that the gender information is not deteriorating the performance, but on the contrary it can be useful towards higher generalization. Indeed, we observe a slight deterioration of the performance for the blondhair and heavymakeup cases. Indeed, this can be observed by the higher difficulty in minimizing the $R_{perp}$ term besides the loss, indicating that the information we desire to keep private is not only learned, but actively used for the classification task.\\
Interestingly, we observe a typical trend in all of our experiments: for the baseline models (or in other words, when $\gamma=0$), the maximum value for the recorded value of $R$, which results also in the maximum private information leakage, is recorded nearby the lowest value of the loss for the validation set. An example of this typical trend we observe in all the conducted experiments is shown in Fig.~\ref{fig::validationtrend}, which shows the trend of R, the loss on the validation set and the loss on the test set for the SFR dataset. Intuitively, we could infer that the information we wish to keep private is necessary to generalize on the given task, and as the model starts over-fitting such information is being obfuscated. However, looking at the performance recorded when applying DisP, for $\gamma\neq 0$, we clearly observe that good generalization performance for low values of R is still possible. This means that in this case there is some natural tendency for the model to implicitly learn this private information, which can be discouraged by applying DisP.\\
For completeness, in Tab.~\ref{tab:cvresults} we also propose the same set of experiments with a 5-fold cross-validation strategy, obtaining similar results as those achieved with the original validation set proposed in CelebA and SFR.
\begin{table*}
    \caption{Results on the trained models with 5-fold cross-validation (in the upper row train set performance, in the bottom test set performance) .}
    \label{tab:cvresults}
    \small
	\renewcommand{\arraystretch}{1.2}
	\centering
		\begin{tabular}{c c c c c c c}
        \toprule
                &       &\multicolumn{2}{c}{\bf Training}&\multicolumn{3}{c}{\bf Attacks}\\
        Dataset & Target & $\gamma$   &Accuracy  & Unsup. & Sup.(1H) & Sup.(2H)\\
                &              &      & [\%]  &  [Train\%-Test\%]  &[Train\%-Test\%]&[Train\%-Test\%]\\
        \midrule
        \multirow{18}{*}{CelebA}
            &\multirow{6}{*}{Blondhair}
                &0           &95.1 $\pm$ 0.1 &62.0 $\pm$ 0.6 &89.8 $\pm$ 0.4 &90.9 $\pm$ 0.3\\
                &&           &               &               &88.1 $\pm$ 0.6 &88.4 $\pm$ 0.3\\\cline{3-7}
                &&0.1        &95.0 $\pm$ 0.1 &58.0 $\pm$ 0.1 &81.2 $\pm$ 0.3 &81.8 $\pm$ 0.3\\
                &&           &               &               &80.5 $\pm$ 1.0 &81.5 $\pm$ 0.6\\\cline{3-7}
                &&0.2        &95.1 $\pm$ 0.2 &58.0 $\pm$ 0.0 &81.3 $\pm$ 1.5 &81.8 $\pm$ 1.5\\
                &&           &               &               &80.6 $\pm$ 1.7 &80.7 $\pm$ 1.2\\
            \cline{2-7}
            &\multirow{6}{*}{Eyeglasses}
                &0           &99.6 $\pm$ 0.0 &54.4$\pm$ 1.3  &87.8 $\pm$ 0.3 &90.2 $\pm$ 0.2\\
                &&           &               &               &85.2 $\pm$ 0.3 &85.2 $\pm$ 0.5\\\cline{3-7}
                &&0.1        &99.7 $\pm$ 0.0 &51.2$\pm$ 0.3  &79.0 $\pm$ 0.8 &79.3 $\pm$ 0.9\\
                &&           &               &               &78.6 $\pm$ 1.1 &78.7 $\pm$ 1.1\\\cline{3-7}
                &&0.2        &99.7 $\pm$ 0.0 &50.7$\pm$ 1.2  &78.3 $\pm$ 0.5 &79.0 $\pm$ 0.5\\
                &&           &               &               &77.6 $\pm$ 0.7 &78.0 $\pm$ 0.5\\
            \cline{2-7}
            &\multirow{6}{*}{Heavymakeup}
                &0           &90.3 $\pm$ 0.0 &86.4$\pm$ 3.8  &94.7 $\pm$ 0.0 &94.7 $\pm$ 0.1\\
                &&           &               &               &94.5 $\pm$ 0.1 &94.4 $\pm$ 0.1\\\cline{3-7}
                &&0.1        &90.1 $\pm$ 0.1 &83.2$\pm$ 0.2  &94.5 $\pm$ 0.2 &94.5 $\pm$ 0.2\\
                &&           &               &               &94.5 $\pm$ 0.2 &94.5 $\pm$ 0.2\\\cline{3-7}
                &&0.2        &90.1 $\pm$ 0.2 &80.4$\pm$ 5.7  &94.7 $\pm$ 0.0 &94.7 $\pm$ 0.2\\
                &&           &               &               &94.6 $\pm$ 0.1 &94.4 $\pm$ 0.2\\
        \midrule
        \multirow{6}{*}{SFR}
            &\multirow{6}{*}{Pneumonia/Typical}
                &0           &77.1$\pm$ 1.0  &77.5$\pm$ 3.7  &85.2 $\pm$ 0.2 &86.2 $\pm$ 0.2\\
                &&           &               &               &82.3 $\pm$ 0.1 &84.9 $\pm$ 0.2\\\cline{3-7}
                &&0.1        &78.2$\pm$ 1.3  &57.2$\pm$ 0.1  &70.4$\pm$ 2.1  &70.4$\pm$ 2.4\\ 
                &&           &               &               &62.6$\pm$ 1.8  &60.5$\pm$ 1.1\\\cline{3-7}
                &&0.2        &77.5$\pm$ 0.8  &58.0$\pm$ 0.1  &67.5$\pm$ 0.5  &69.1$\pm$ 3.5\\
                &&           &               &               &60.0$\pm$ 1.5  &58.7$\pm$ 1.4\\
        \bottomrule
	\end{tabular}
\end{table*}
\subsection{Attacks}
Looking at the unsupervised attacks, we interestingly observe that there are certain scenarios in which the information is easier to extract. For example, when the target is Blondhair in CelebA, the unsupervised attack is able to recover with the 73\% the private information, evidencing a problem of information leakage. Such a problem is even more evident with the SFR dataset, where an unsupervised attack recovers the private features with the 75\% accuracy. Of course, there are scenarios in which the private feature is not learned and, for instance, it is harder to extract: it ts the case for the Eyeglasses feature, in the CelebA dataset. Intuitively, such a feature is naturally de-correlated with the gender, and for instance such an information is unlikely to leak from the bottleneck layer.\\
Looking at the supervised attack, we observe that the performance is typically higher than the unsupervised attack, as expected. While for the CelebA-Eyeglasses scenario the classifier is (almost) random guessing for all our experiments, that is not the case for the other scenarios. Focusing on the baseline models ($\gamma=0$), we observe that the generalization performance on the test set shows the tendency to improve by adding more complexity to the classifier (comparing H1 to H2). However, using DisP, we observe a decreasing generalization trend when increasing the classifier complexity. We have not used any validation/fine-tuning strategy to train the supervised attack: we have used SGD with fixed learning rate of 0.1, batch size 100, weight decay 0.9 for 50 epochs, for all the experiments. Under these conditions, since the train accuracy increases but the test accuracy decreases, we can say the classifier is driven in a memorization state: this means that it is simply unable to extract general features to learn the private classes, as they are disentangled. As the value of $\gamma$ increases (meaning that the weight of our regularization term in the learning process increases), the performance achieved by the attacks decreases.

\subsection{Applicability beyond deep learning}
DisP proposes itself as a regularization strategy towards a disentanglement of private classes, and as such it is implicitly bound to gradient-based optimization techniques. However, its applicability is not necessarily limited to deep learning models: other machine learning algorithms are compatible with DisP, like for example Support Vector Machines (SVMs), where evidently the impact of DisP will be on the class' separator. Besides, DisP can virtually be applied isolatedly as a perceptron besides any other machine learning strategy (like for example, in a compression pipeline after PCA).

%% file: sections/6_conclusion.tex
\section{Conclusion}
\label{sec:conclusion}
In this work we have proposed DisP, a regularization term which disentangles private classes. In particular, we favor the selection of features correlating with the target class filtering the propagation of features correlating with the private classes.In order to be more interdisciplinary and to create a more robust and ethical AI technology, the private features selection was made by keeping some key concepts of the GDPR in mind. The GDPR is an important legislative act whose objective is to protect people’s privacy not only through regulatory and compliance functions but also through principles such as the one of privacy by default and by design, which is bound to influence emerging technologies, AI included. Once the features to be considered ``private'' are detected, DisP is able to disentangle them at some bottleneck layer, minimizing the risk of extraction for this information. We have shown the effectiveness of the models trained with DisP attempting both unsupervised and supervised attacks to retrieve the information. In particular, we have tested on face images and CXR images, showing that in certain cases there is unwanted information leakage it can be successfully recovered, unless the models are not DisP regularized. Future work include the development of a layer able to handle quantized features on top of which mutual information can be directly measured and minimized.

%% file: main.bbl
\begin{thebibliography}{10}
\providecommand{\url}[1]{#1}
\csname url@samestyle\endcsname
\providecommand{\newblock}{\relax}
\providecommand{\bibinfo}[2]{#2}
\providecommand{\BIBentrySTDinterwordspacing}{\spaceskip=0pt\relax}
\providecommand{\BIBentryALTinterwordstretchfactor}{4}
\providecommand{\BIBentryALTinterwordspacing}{\spaceskip=\fontdimen2\font plus
\BIBentryALTinterwordstretchfactor\fontdimen3\font minus
  \fontdimen4\font\relax}
\providecommand{\BIBforeignlanguage}[2]{{%
\expandafter\ifx\csname l@#1\endcsname\relax
\typeout{** WARNING: IEEEtran.bst: No hyphenation pattern has been}%
\typeout{** loaded for the language `#1'. Using the pattern for}%
\typeout{** the default language instead.}%
\else
\language=\csname l@#1\endcsname
\fi
#2}}
\providecommand{\BIBdecl}{\relax}
\BIBdecl

\bibitem{sonoda2017neural}
S.~Sonoda and N.~Murata, ``Neural network with unbounded activation functions
  is universal approximator,'' \emph{Applied and Computational Harmonic
  Analysis}, vol.~43, no.~2, pp. 233--268, 2017.

\bibitem{selvaraju2017grad}
R.~R. Selvaraju, M.~Cogswell, A.~Das, R.~Vedantam, D.~Parikh, and D.~Batra,
  ``Grad-cam: Visual explanations from deep networks via gradient-based
  localization,'' in \emph{Proceedings of the IEEE international conference on
  computer vision}, 2017, pp. 618--626.

\bibitem{Osia2020AHD}
S.~A. Osia, A.~S. Shamsabadi, S.~Sajadmanesh, A.~Taheri, K.~Katevas, H.~Rabiee,
  N.~Lane, and H.~Haddadi, ``A hybrid deep learning architecture for
  privacy-preserving mobile analytics,'' \emph{IEEE Internet of Things
  Journal}, vol.~7, pp. 4505--4518, 2020.

\bibitem{art1}
``Regulation (eu) 2016/679 of the european parliament and of the council of 27
  april 2016 on the protection of natural persons with regard to the processing
  of personal data and on the free movement of such data, and repealing
  directive 95/46/ec (general data protection regulation),'' \emph{OJ}, vol. L
  119, pp. 1--88, 4.5.2016.

\bibitem{art2}
``Ethics guidelines for trustworthy ai.''
  \url{https://digital-strategy.ec.europa.eu/en/library/ethics-guidelines-trustworthy-ai}.

\bibitem{art3}
``Assessment list for trustworthy artificial intelligence (altai) for
  self-assessment.''
  \url{https://digital-strategy.ec.europa.eu/en/library/assessment-list-trustworthy-artificial-intelligence-altai-self-assessment}.

\bibitem{art4}
``Proposal for a regulation of the european parliament and of the council
  laying down harmonised rules on artificial intelligence (artificial
  intelligence act) and amending certain union legislative acts com/2021/206
  final.''

\bibitem{purtova2018law}
N.~Purtova, ``The law of everything. broad concept of personal data and future
  of eu data protection law,'' \emph{Law, Innovation and Technology}, vol.~10,
  no.~1, pp. 40--81, 2018.

\bibitem{leenes2018artificial}
R.~Leenes and S.~De~Conca, ``Artificial intelligence and privacy—ai enters
  the house through the cloud,'' in \emph{Research handbook on the law of
  artificial intelligence}.\hskip 1em plus 0.5em minus 0.4em\relax Edward Elgar
  Publishing, 2018.

\bibitem{voigt2017eu}
P.~Voigt and A.~Von~dem Bussche, ``The eu general data protection regulation
  (gdpr),'' \emph{A Practical Guide, 1st Ed., Cham: Springer International
  Publishing}, vol.~10, p. 3152676, 2017.

\bibitem{bygrave2017data}
L.~A. Bygrave, ``Data protection by design and by default: deciphering the
  eu’s legislative requirements,'' \emph{Oslo Law Review}, vol.~4, no.~02,
  pp. 105--120, 2017.

\bibitem{horvath2015attribute}
M.~Horv{\'a}th, ``Attribute-based encryption optimized for cloud computing,''
  in \emph{International Conference on Current Trends in Theory and Practice of
  Informatics}.\hskip 1em plus 0.5em minus 0.4em\relax Springer, 2015, pp.
  566--577.

\bibitem{zhang2020attribute}
Y.~Zhang, R.~H. Deng, S.~Xu, J.~Sun, Q.~Li, and D.~Zheng, ``Attribute-based
  encryption for cloud computing access control: A survey,'' \emph{ACM
  Computing Surveys (CSUR)}, vol.~53, no.~4, pp. 1--41, 2020.

\bibitem{singh2020probabilistic}
A.~Singh, S.~Garg, R.~Kaur, S.~Batra, N.~Kumar, and A.~Y. Zomaya,
  ``Probabilistic data structures for big data analytics: A comprehensive
  review,'' \emph{Knowledge-Based Systems}, vol. 188, p. 104987, 2020.

\bibitem{acar2018survey}
A.~Acar, H.~Aksu, A.~S. Uluagac, and M.~Conti, ``A survey on homomorphic
  encryption schemes: Theory and implementation,'' \emph{ACM Computing Surveys
  (Csur)}, vol.~51, no.~4, pp. 1--35, 2018.

\bibitem{warner1965randomized}
S.~L. Warner, ``Randomized response: A survey technique for eliminating evasive
  answer bias,'' \emph{Journal of the American Statistical Association},
  vol.~60, no. 309, pp. 63--69, 1965.

\bibitem{fellegi1972question}
I.~P. Fellegi, ``On the question of statistical confidentiality,''
  \emph{Journal of the American Statistical Association}, vol.~67, no. 337, pp.
  7--18, 1972.

\bibitem{dwork2016calibrating}
C.~Dwork, F.~McSherry, K.~Nissim, and A.~Smith, ``Calibrating noise to
  sensitivity in private data analysis,'' \emph{Journal of Privacy and
  Confidentiality}, vol.~7, no.~3, pp. 17--51, 2016.

\bibitem{dwork2009differential}
C.~Dwork and J.~Lei, ``Differential privacy and robust statistics,'' in
  \emph{Proceedings of the forty-first annual ACM symposium on Theory of
  computing}, 2009, pp. 371--380.

\bibitem{duchi2014privacy}
J.~C. Duchi, M.~I. Jordan, and M.~J. Wainwright, ``Privacy aware learning,''
  \emph{Journal of the ACM (JACM)}, vol.~61, no.~6, pp. 1--57, 2014.

\bibitem{abadi2016deep}
M.~Abadi, A.~Chu, I.~Goodfellow, H.~B. McMahan, I.~Mironov, K.~Talwar, and
  L.~Zhang, ``Deep learning with differential privacy,'' in \emph{Proceedings
  of the 2016 ACM SIGSAC conference on computer and communications security},
  2016, pp. 308--318.

\bibitem{chamikara2019efficient}
M.~A.~P. Chamikara, P.~Bert{\'o}k, D.~Liu, S.~Camtepe, and I.~Khalil, ``An
  efficient and scalable privacy preserving algorithm for big data and data
  streams,'' \emph{Computers \& Security}, vol.~87, p. 101570, 2019.

\bibitem{shokri2015privacy}
R.~Shokri and V.~Shmatikov, ``Privacy-preserving deep learning,'' in
  \emph{Proceedings of the 22nd ACM SIGSAC conference on computer and
  communications security}, 2015, pp. 1310--1321.

\bibitem{He2016DeepRL}
K.~He, X.~Zhang, S.~Ren, and J.~Sun, ``Deep residual learning for image
  recognition,'' \emph{2016 IEEE Conference on Computer Vision and Pattern
  Recognition (CVPR)}, pp. 770--778, 2016.

\bibitem{Huang2017DenselyCC}
G.~Huang, Z.~Liu, and K.~Q. Weinberger, ``Densely connected convolutional
  networks,'' \emph{2017 IEEE Conference on Computer Vision and Pattern
  Recognition (CVPR)}, pp. 2261--2269, 2017.

\bibitem{Ronneberger2015UNetCN}
O.~Ronneberger, P.~Fischer, and T.~Brox, ``U-net: Convolutional networks for
  biomedical image segmentation,'' in \emph{MICCAI}, 2015.

\bibitem{Tishby2015DeepLA}
N.~Tishby and N.~Zaslavsky, ``Deep learning and the information bottleneck
  principle,'' \emph{2015 IEEE Information Theory Workshop (ITW)}, pp. 1--5,
  2015.

\bibitem{Goldfeld2020TheIB}
Z.~Goldfeld and Y.~Polyanskiy, ``The information bottleneck problem and its
  applications in machine learning,'' \emph{IEEE Journal on Selected Areas in
  Information Theory}, vol.~1, pp. 19--38, 2020.

\bibitem{tartaglione2021end}
E.~Tartaglione, C.~A. Barbano, and M.~Grangetto, ``End: Entangling and
  disentangling deep representations for bias correction,'' in
  \emph{Proceedings of the IEEE/CVF Conference on Computer Vision and Pattern
  Recognition}, 2021, pp. 13\,508--13\,517.

\bibitem{Tartaglione2020ANA}
E.~Tartaglione and M.~Grangetto, ``A non-discriminatory approach to ethical
  deep learning,'' \emph{2020 IEEE 19th International Conference on Trust,
  Security and Privacy in Computing and Communications (TrustCom)}, pp.
  943--950, 2020.

\bibitem{buchta2012spherical}
C.~Buchta, M.~Kober, I.~Feinerer, and K.~Hornik, ``Spherical k-means
  clustering,'' \emph{Journal of statistical software}, vol.~50, no.~10, pp.
  1--22, 2012.

\bibitem{bahng2019rebias}
H.~Bahng, S.~Chun, S.~Yun, J.~Choo, and S.~J. Oh, ``Learning de-biased
  representations with biased representations,'' in \emph{International
  Conference on Machine Learning (ICML)}, 2020.

\bibitem{lecun2010mnist}
Y.~LeCun, C.~Cortes, and C.~Burges, ``Mnist handwritten digit database,''
  \emph{ATT Labs [Online]. Available: http://yann.lecun.com/exdb/mnist},
  vol.~2, 2010.

\bibitem{liu2015faceattributes}
Z.~Liu, P.~Luo, X.~Wang, and X.~Tang, ``Deep learning face attributes in the
  wild,'' in \emph{Proceedings of International Conference on Computer Vision
  (ICCV)}, December 2015.

\end{thebibliography}
